%% file: CIKM_main.tex
\documentclass[sigconf]{acmart}
\usepackage[utf8]{inputenc} 
\usepackage[T1]{fontenc}    
\usepackage{hyperref}       
\usepackage{url}            
\usepackage{booktabs}       
\usepackage{amsfonts}       
\usepackage{nicefrac}       
\usepackage{microtype}      
\usepackage{lipsum}
\usepackage{fancyhdr}       
\usepackage{graphicx}       

\usepackage{todonotes}
\usepackage{amsmath}
\usepackage{algorithmic}
\usepackage{textcomp}
\usepackage{xcolor}
\usepackage{multirow}
\usepackage{natbib}
\usepackage{wrapfig}
\usepackage{dsfont}

\input{macros}

\input{math_commands}

\AtBeginDocument{%
  }

\setcopyright{acmlicensed}
\copyrightyear{2025}
\acmYear{2025}
\acmDOI{10.1145/3746252.3761341}
\acmConference[CIKM '25] {Proceedings of the 34th ACM International Conference on Information and Knowledge Management}{ November 10--14, 2025}{Seoul, Republic of Korea.}
\acmBooktitle{Proceedings of the 34th ACM International Conference on Information and Knowledge Management (CIKM '25), November 10--14, 2025, Seoul, Republic of Korea}
\acmISBN{979-8-4007-2040-6/2025/11}
\acmDOI{10.1145/3746252.3761341}
\acmISBN{978-1-4503-XXXX-X/2018/06}

\acmSubmissionID{8159}

\begin{document}

\title{FROG: Fair Removal on Graph}

\author{Ziheng Chen}
\email{albertchen1993pokemon@gmail.com}
\orcid{0000-0002-2585-637X}
\affiliation{%
  \institution{Walmart Global Tech, USA}
  \country{}
}

\author{Jiali Cheng}
\email{jiali\_cheng@uml.edu}
\affiliation{%
  \institution{University of Massachusetts Lowell, USA}
  \country{}
}

\author{Hadi Amiri}
\email{Hadi_Amiri@uml.edu}
\affiliation{%
  \institution{University of Massachusetts Lowell, USA}
  \country{}
}

\author{Kaushiki Nag}
\email{Kaushiki.Nag@walmart.com}
\affiliation{%
  \institution{Walmart Global Tech, USA}
  \country{}
}

\author{Lu Lin}
\email{lulin@psu.edu}
\affiliation{%
  \institution{Pennsylvania State University, USA}
  \country{}
}

\author{Sijia Liu}
\email{liusiji5@msu.edu}
\affiliation{%
  \institution{Michigan State University, USA}
  \country{}
}

\author{Xiangguo Sun}
\email{xiangguosun@cuhk.edu.hk}
\affiliation{%
  \institution{The Chinese University of Hong Kong, China}
  \country{}
}

\author{Gabriele Tolomei}
\email{tolomei@di.uniroma1.it}
\affiliation{%
  \institution{Sapienza University of Rome, Italy}
  \country{}
}

\renewcommand{\shortauthors}{Chen et al.}


\input{000abstract}




\maketitle


\input{010intro}
\input{030preliminary}

\input{040reason}
\input{050formulation}

\input{060method}

\input{070experiments}
\input{080results}
\input{020related}
\input{090conclusion}

\begin{acks}
This research was, in part, funded by the National Institutes of Health (NIH) Agreement No. 1OT2OD032581. The views and conclusions contained in this document are those of the authors and should not be interpreted as representing the official policies, either expressed or implied, of the NIH.
This work was also partially supported by the following projects:
SERICS (PE00000014) under the National Recovery and Resilience Plan funded by the European Union NextGenerationEU; HyperKG -- Hybrid Prediction and Explanation with Knowledge Graphs (2022Y34XNM) funded by the Italian Ministry of University and Research under the PRIN 2022 program; GHOST -- Protecting User Privacy from Community Detection in Social Networks (B83C24007070005) funded by Sapienza University of Rome - ``Progetti di Ricerca Grandi.'' 
\end{acks}

\appendix
\input{091appendix}

\section*{GenAI Disclosure Statement}
We used GPT-4 to identify and correct grammatical errors, typos, and to improve the overall writing quality. 
No AI tools were used at any other stage of this work to ensure full academic integrity.
\bibliographystyle{ACM-Reference-Format}
\balance
\bibliography{mybibfile}



\end{document}

%% file: macros.tex
\usepackage{xspace}

\newcommand{\method}{\textsc{FROG}\xspace}

\newcommand{\V}{\mathcal{V}}

\newcommand{\inputs}{\mathcal{V}}
\newcommand{\outputs}{\mathcal{Y}}

\newcommand{\weights}{\boldsymbol{\omega}}

\newcommand{\ori}{g_{\weights}}
\newcommand{\un}{g_{u}}
\newcommand{\optimal}{g_{\weights^{*}}}

\newcommand{\graph}{\mathcal{G}}

\newcommand{\edge}{\mathcal{E}}

%% file: math_commands.tex
\usepackage{amsmath,amsfonts,bm}

\def\eqref#1{equation~\ref{#1}}

\def\1{\bm{1}}

\def\vz{{\bm{z}}}

\DeclareMathAlphabet{\mathsfit}{\encodingdefault}{\sfdefault}{m}{sl}
\SetMathAlphabet{\mathsfit}{bold}{\encodingdefault}{\sfdefault}{bx}{n}



%% file: 000abstract.tex
\begin{abstract}
With growing emphasis on privacy regulations, \textit{machine unlearning} has become increasingly critical in real-world applications such as social networks and recommender systems, many of which are naturally represented as graphs. However, existing graph unlearning methods often modify nodes or edges indiscriminately, overlooking their impact on fairness. For instance, forgetting links between users of different genders may inadvertently exacerbate group disparities. 
To address this issue, we propose a novel framework that jointly optimizes both the graph structure and the model to achieve \textit{fair} unlearning. Our method rewires the graph by removing redundant edges that hinder forgetting while preserving fairness through targeted edge augmentation. We further introduce a worst-case evaluation mechanism to assess robustness under challenging scenarios. Experiments on real-world datasets show that our approach achieves more effective and fair unlearning than existing baselines.

\end{abstract}

%% file: 010intro.tex
\section{Introduction}
\label{sec:intro}

Recent breakthroughs in deep learning have significantly advanced artificial intelligence (AI) systems across various domains. In particular, graph neural networks (GNNs) have emerged as a standard approach for addressing graph-related tasks~\cite{wang2024self,zhang2023mixupexplainer}, such as node and edge classification -- fundamental for applications in social networks (e.g., friend recommendations) and biochemistry (e.g., drug discovery). 
However, the widespread adoption of GNNs raises concerns about privacy leakage, as training data containing sensitive relationships can be implicitly ``memorized'' within model parameters. To mitigate the risk of misuse, recent regulatory policies have established the \textit{right to be forgotten} \cite{TRTBF2013}, allowing users to remove private data from online platforms. Consequently, a range of graph unlearning methods have been developed to effectively erase specific knowledge from trained GNNs without requiring full retraining.

Although graph unlearning effectively removes edges/nodes, its potential risks -- particularly \textit{disparate impact} -- are often overlooked. In link prediction, disparate impact refers to disparities in links that stem from sensitive attributes such as gender or race, which are protected under anti-discrimination laws. Recent studies suggest that changes in graph topology, characterized by homophily ratios (see Section~\ref{sec:reason}), can exacerbate bias through feature propagation. For instance, in social networks, removing links to opposite-sex friends may lead to an increased likelihood of users being recommended connections within the same gender group. Thus, long-term accumulation could result in social segregation.

\begin{figure}
\centering
    \includegraphics[width=0.6\linewidth]{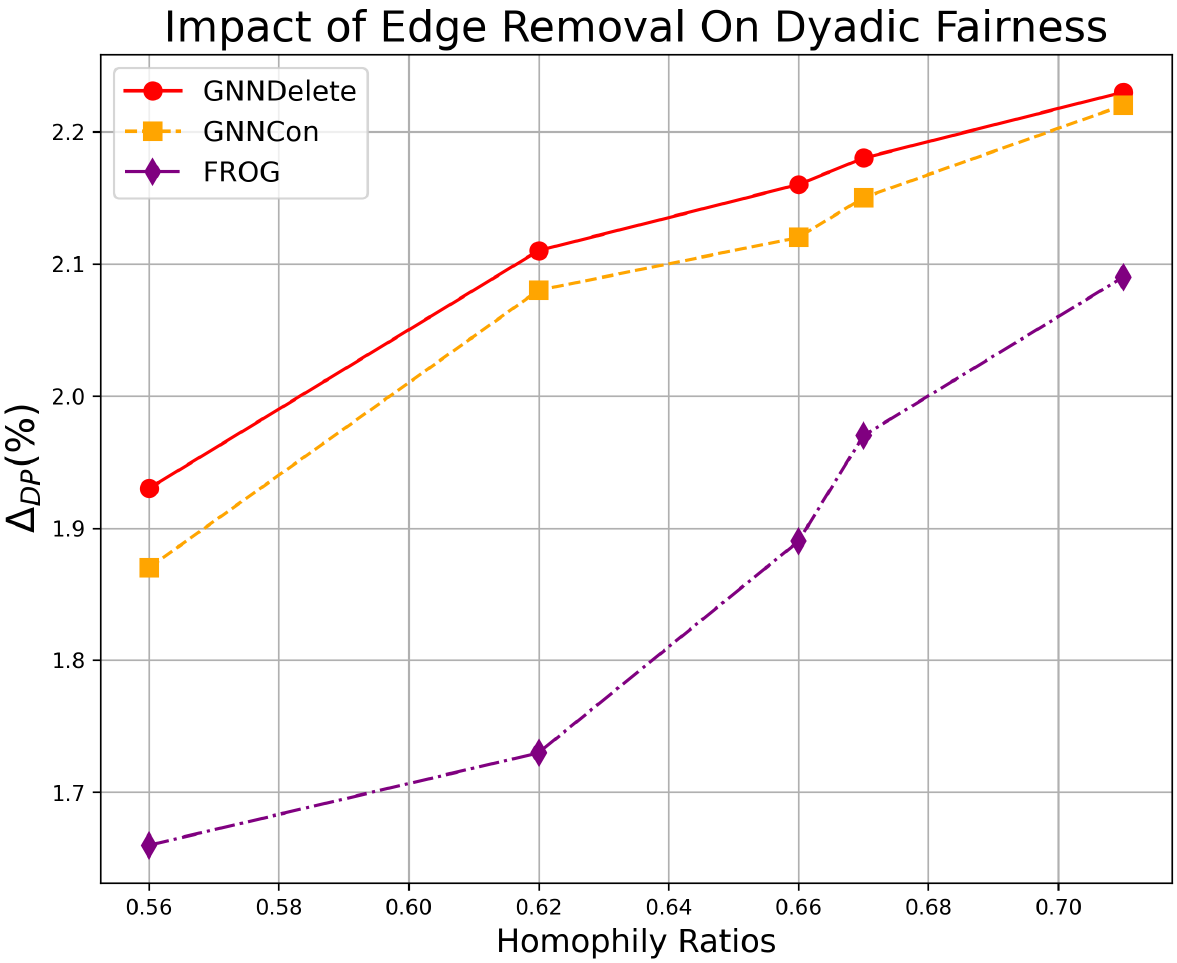}
    \caption{The impact of removing edges on the fairness of graph unlearning algorithms is shown. The $x$-axis represents the homophily ratio, defined as the proportion of a node’s neighbors with identical sensitive features. And the $y$-axis indicates $\Delta_{DP}$, a measure of dyadic fairness.}

    \label{fig:impact}
\end{figure}


Recently, several algorithms have achieved strong performance in graph unlearning. However, we observed a significant impact on fairness, as edge removal requests alter the graph topology. To examine this effect in social networks, we evaluate two state-of-the-art methods, GNNDelete~\cite{cheng2023gnndelete} and GNNCon~\cite{yang2023contrastive}, on Facebook$\#1684$~\cite{li2021dyadic}, a social ego network from Facebook app, using gender as the sensitive feature. As shown in Figure~\ref{fig:impact}, both methods fail to maintain dyadic fairness, measured by $\Delta_{DP}$ (see Section~\ref{sec:reason}), when increasing edge removal requests lead to a higher homophily ratio.


The underlying reason is that current algorithms focus solely on designing loss functions to reduce the prediction probability of forgotten edges, without accounting for the bias introduced by edge removal. Moreover, they have also been criticized for \textit{under-forgetting}~\cite{cheng2023gnndelete}, where an algorithm fails to forget certain edges even after sufficient epochs of unlearning. Consequently, we argue that existing unlearning algorithms do not fully leverage the potential of the graph structure and may not achieve optimal performance.

In this paper, we study a novel and detrimental phenomenon where existing unlearning algorithms can alter the graph structure, inadvertently introducing bias. To address this issue, we propose \textbf{FROG} (\textit{Fair Removal on Graph}), a framework designed to effectively forget target knowledge while simultaneously mitigating disparate impact. Our key contributions are as follows:


\begin{itemize}
\item \textbf{Problem:} We present the first investigation on how graph unlearning impacts graph homophily and disrupts node embeddings through the aggregation mechanism in GNNs, potentially exacerbating discrimination in downstream tasks.
\item \textbf{Algorithm:} We propose a novel framework, for fair graph unlearning, which integrates graph rewiring and model updating. The graph is rewired by adding edges to mitigate the bias introduced by the deletion request, while removing redundant edges that hinder unlearning. Furthermore, the framework is adaptable to any graph-based unlearning methods for model updates. 
\item \textbf{Evaluation:} In order to truly gauge the authenticity of unlearning performance, we introduce the concept of the ``worst-case forget set'' in graph unlearning. Experiments on real-world datasets demonstrate that our method improves unlearning effectiveness while mitigating discrimination.
\end{itemize}


%% file: 030preliminary.tex
\section{Preliminaries}

\noindent \textbf{\textit{Graph Neural Networks.}} 
We consider an undirected attributed graph $\graph=(\V,\edge, \mathbf{X})$ with nodes set $\V$, edge set $\edge$ and node features $\mathbf{X}$.
Each node is also associated with a categorical sensitive attribute $s_i \in S$ (e.g., political preference, gender), which may or may not be part of its features. The graph topology can be summarized by the adjacency matrix $\mathbf{A}$. Also, we introduce a predictive Graph Neural Network (GNN) model $\ori: \inputs \mapsto \outputs$, with parameters $\weights$, to predict the nodes' labels as follows: 
\[\hat{Y}=f(\mathbf{Z}), \quad \mbox{with}\quad \mathbf{Z}=\ori(\mathbf{X},\mathbf{A}),\]
where $\mathbf{Z}$ represents the node embedding and $\hat{Y}\in \outputs$ is the predicted label. The dot product between node embeddings $\mathbf{z}_i^T \mathbf{z}_j$ is used to predict whether edge $e_{ij}$ exists. Also, we refer to $\ori$ as the ``original model'' prior to unlearning. \\

\noindent \textbf{\textit{Graph Unlearning.}}
Graph unlearning involves selectively removing certain instances or knowledge from a trained model without the need for full retraining. Given a graph $\mathcal{G} = (\mathcal{V}, \mathcal{E}, \mathbf{X})$
and a subset of its elements $\mathcal{G}_f = (\mathcal{V}_f,\mathcal{E}_f, \mathbf{X}_f)$ to be unlearned, we denote the retained subgraph as $\mathcal{G}_r = (\mathcal{V}_r, \mathcal{E}_r, \mathbf{X}_r)$, where $\mathcal{G}_r = \mathcal{G} \setminus \mathcal{G}_f$,
with the conditions $\mathcal{G}_f \cup \mathcal{G}_r = \mathcal{G}$ and $\mathcal{G}_f \cap \mathcal{G}_r = \emptyset.$
Graph unlearning aims to obtain an unlearned model, denoted as $\un$, that behaves as if it were trained solely on $\mathcal{G}_r$. Requests for graph unlearning can be broadly categorized into two types:
\textit{edge deletion}, where a subset $\mathcal{E}_f \subset \mathcal{E}$ is removed, and
\textit{node deletion}, where a subset $\mathcal{V}_f \subset \mathcal{V}$ is removed.

The goal is to derive a new model $\un$ from the original model $\ori$ that no longer contains the information from $\mathcal{G}_f$, while preserving its performance on $\mathcal{G}_r$. Since fully retraining the model on $\mathcal{G}_r$ to obtain an optimal model, denoted $\optimal$, is often time-consuming, our objective is to approximate $\optimal$ by updating $\ori$ using the unlearning process based on $\mathcal{G}_f$ as follows:
\begin{equation*}
    \ori \xrightarrow{\mathcal{G}_f} \un \approx \optimal.
\end{equation*}

\noindent \textbf{\textit{Fairness for Graph Data.}}
In this work, we focus on group fairness (also known as disparate impact), which emphasizes that algorithms should not yield discriminatory outcomes for any specific demographic group~\cite{wang2022improving}.

For example, in the node classification task, group fairness aims to mitigate the influence of sensitive attributes on individual predictions.  Assuming both the target outcome and $S$ are binary-valued, a widely used criterion is \textit{Demographic Parity} (DP)~\cite{spinelli2021fairdrop}: a classifier satisfies DP if the likelihood of a positive outcome is the same regardless of the value of the sensitive attribute $S$:
\[P(\hat{Y}|S=1)=P(\hat{Y}|S=0).\]
In link prediction, we consider the disparity in link formation between intra- and inter-sensitive groups. Extending from \textit{Demographic Parity}, we adopt \textit{Dyadic Fairness }~\cite{li2021dyadic}, which requires that the predicted likelihood of a link is independent of whether the connected nodes share the same sensitive attribute. A link prediction algorithm satisfies \textit{Dyadic Fairness} if its predictive scores meet the following condition:
\[P(g(u,v)|S(u)=S(v))=P(g(u,v)|S(u)\neq S(v)).\]
Here, we assume the link prediction function $g$ is modeled as the inner product of nodes' embeddings.

%% file: 040reason.tex
\section{Motivation}\label{sec:reason}
In this section, we present a series of theoretical analyses to elucidate how graph unlearning can lead to unfairness. During the unlearning process, the removal of edges may exacerbate the network homophily, where nodes with similar
sensitive features tend to form closer connections than dissimilar ones, inevitably disrupting information flow of graph neural network between nodes within and across sensitive groups.


Inspired by previous work~\cite{wang2022improving,cui2023event},
we reveal how the node homophily ratio $\rho$, which is defined as the proportion of a node's neighbors sharing the same sensitive features as the node, can amplify the bias. For simplicity, we focus on a single-layer graph neural network model.
 Without loss of generality, we assume that node features from two sensitive groups in the network independently and identically follow two different Gaussian distribution $\mathbf{X}^{S_0}\sim \mathcal{N}(\mu^{0},\Sigma^{0}),\mathbf{X}^{S_1}\sim \mathcal{N}(\mu^{1},\Sigma^{1})$. We proved this theorem in the Appendix.

\begin{theorem}
    Given a 1-layer $\ori$ with row-normalized
adjacency $\tilde{\mathbf{A}}=\mathbf{D}^{-1}\mathbf{A}$ ($\mathbf{D}$ is the degree matrix) for feature smoothing and weight matrix $\mathbf{W}$. Suppose $\exists K>0, \forall v\in \mathcal{V}, ||\mathbf{X}_v||_2\leq K$, then the dyadic fairness follows:
\begin{equation}
\Delta_{DP}=|E_{\substack{(v,u)\\ S_{u}= S_{v}}}[\mathbf{z}_v\cdot \mathbf{z}_u]-E_{\substack{(v,u)\\ S_{u}\neq S_{v}}}[\mathbf{z}_v \cdot \mathbf{z}_u]|\leq |K\cdot(2\rho-1)\mathbf{W}\delta|,
\end{equation}
\end{theorem}
\noindent where $\delta=\mu^{0}-\mu^{1}$, and $\rho$ denotes the homophily ratio defined as: \[\rho=E_{v\in \mathcal{V}}\frac{|\sum_{u\in N(v)}\mathds{1}\{S(v)=S(u)\}|}{|N(v)|}.\]

Theorem 3.1 shows that dyadic fairness is bounded by network homophily $\rho$.  As $\rho$ increases, due to edge removal requests between nodes with different sensitive attributes, $\Delta_{DP}$ may get enlarged. Conversely, decreasing $\rho$ by adding edges between such nodes enhances cross-group neighborhood connectivity. This smoothing effect on node representations helps mitigate bias. The theoretical findings motivate our algorithmic design presented in next section.\looseness-1

%% file: 050formulation.tex
\begin{figure*}[t]
    \centering
    \includegraphics[width=0.9\linewidth]{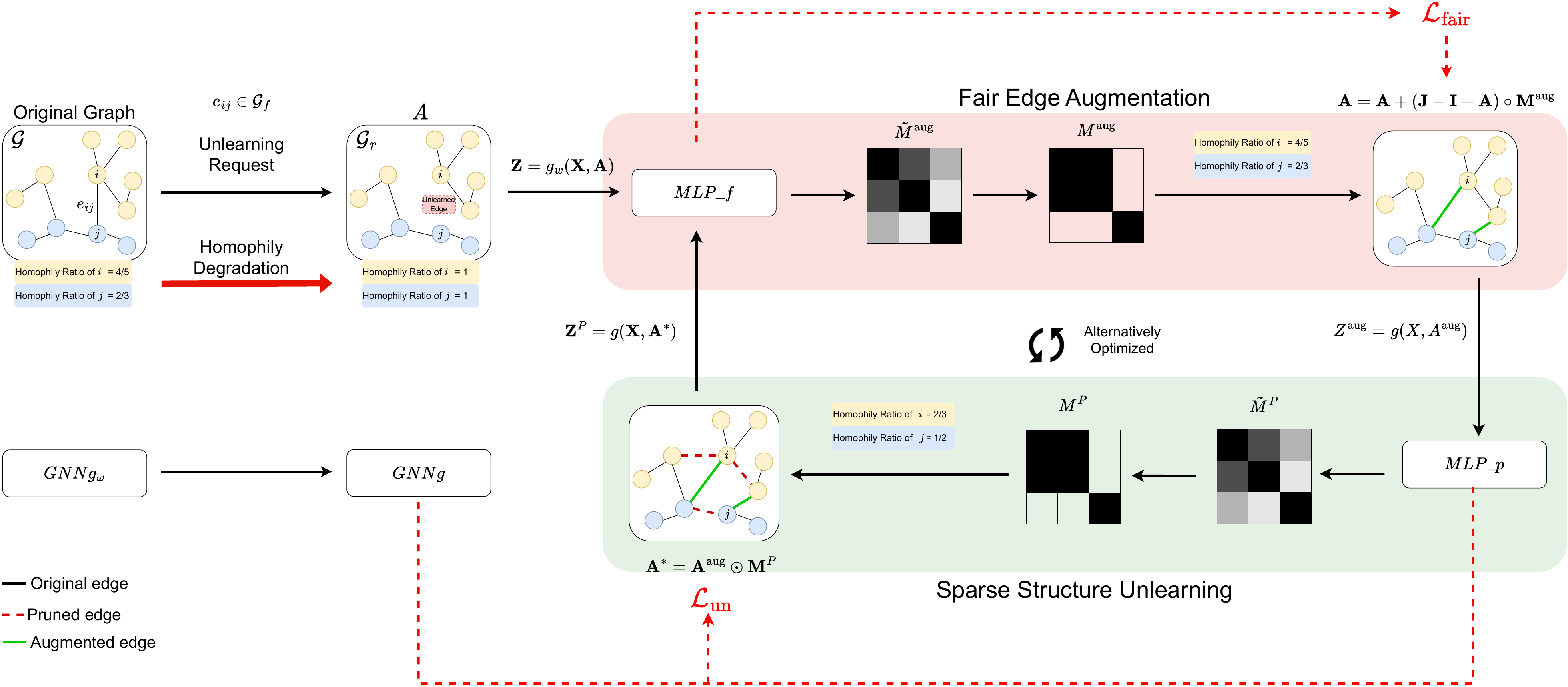}
    \caption{Schematics of \method: $\bm{(i)}$ Request for edges removal; $\bm{(ii)}$ Adding edges to mitigate bias caused by edge removal; and $\bm{(iii)}$ Removing redundant edges that obstruct the unlearning process.}
    \label{fig:method}
\end{figure*}


\section{Problem Formulation}


Given a graph $\graph=(\V,\edge, \mathbf{X})$ and a fully trained GNN $\ori$, our goal is to unlearn each edge $e_{uv} \in \mathcal{E}_f$ from $\ori$, where $\mathcal{E}_f$ denotes the edges to be removed, while mitigating the bias introduced by this removal. Note that node removal can be interpreted as removing all connected edges. The graph structure $\edge$ and the GNN $\ori$ together determine the node embeddings $\mathbf{Z}$, which in turn affect both prediction accuracy and representation fairness. There is broad evidence in literature that graph topology has fundamental effect on the representation~\cite{wang2022improving, li2021dyadic}. Therefore, we aim to simultaneously obtain an unlearned model $\un$ and an optimal graph structure $\mathbf{A}^{*}$. More formally, the task of fair graph unlearning can be cast as:
\begin{equation}
\label{eq:Main Goal}
\begin{aligned}
\un, \mathbf{A^{*}} &= \mbox{arg} \min\limits_{g,\mathbf{\hat{A}}}\mathcal{L}_{\mathrm{un}}(g, \mathbf{\hat{A}}, \mathbf{X}) + \alpha  \mathcal{L}_{\mathrm{fair}}(g, \mathbf{\hat{A}}, \mathbf{X}) \\ 
\textrm{subject to: } & ||\mathbf{A}-\mathbf{A^{*}}||_{0}\leq N. \\
\end{aligned}
\end{equation}
The first component ($\mathcal{L}_{\mathrm{un}}$) is an unlearning loss designed to reduce the memorization of forgetting edges $\edge_f$, while preserving performance on $\edge_r$. Notably, our approach is built to integrate seamlessly with any graph-based unlearning method, allowing any differentiable unlearning loss $\mathcal{L}_{\mathrm{un}}$ to be incorporated in a plug-and-play manner. The second component ($\mathcal{L}_{\mathrm{fair}}$) penalizes violations of representation fairness. In addition, $\alpha$ serves as a scaling factor to trade off between $\mathcal{L}_{\mathrm{un}}$ and $\mathcal{L}_{\mathrm{fair}}$. The detailed form of these losses will be introduced in the following section. 

Besides, to avoid omitting  much information from $\mathbf{A}$, we discourages $\mathbf{A^{*}}$ to be too far away from $\mathbf{A}$ by limiting the number of edges to be modified, i.e., to a maximum of $N$ edges. Here we adopt $L^{0}$-norm to quantify the distance between $\mathbf{A^{*}}$ and $\mathbf{A}$.

%% file: 060method.tex
\section{\method: Fair Removal on Graph}

Directly solving Equation~\ref{eq:Main Goal} is a non-trivial task, particularly when $\mathcal{L}_{\text{un}}$ and $\mathcal{L}_{\text{fair}}$ are in conflict with each other, leading to an imbalance of both objectives. 
To address this, we formulate the process as a Stackelberg game~\citep{von1953theory}, or leader-follower game. First, the leader performs \emph{fair edge augmentation} to recover the fairness degradation caused by unlearning requests. Then, the follower performs \emph{sparse structure unlearning} to achieve the unlearning objective. By alternatively optimizing these two objectives, we aim at finding a balance between preservation of fairness and effective unlearning.
In addition to the GNN parameters, the underlying graph topology is also being optimized, since graph topology has a fundamental effect on graph fairness (see Section~\ref{sec:reason}). 


%
In the upper problem of \emph{fair edge augmentation} taken by the leader, an augmenter $f$ takes an input $\mathbf{A}$ and produces an augmented graph $\mathbf{A}^{\text{aug}}=f(\mathbf{A})$. 
In the lower problem of \emph{sparse structure unlearning} taken by the follower, a pruner $p$ removes redundant edges from $\mathbf{A}^{\text{aug}}$ to get the optimal graph $\mathbf{A}^{*}=p(\mathbf{A}^{\text{aug}})$.
As shown in Figure~\ref{fig:method}, leveraging edge augmentation to explore fair structures, subsequently refined through pruning to align with unlearning objectives, increases the potential to escape sub-optimal solutions. These iterative steps can be unified by formulating the problem as the following bi-level optimization:

\begin{equation}
\small
\label{eq: Two-step Goal}
\begin{aligned}
\un, p=&\mbox{arg} \min\limits_{g, p}\mathcal{L}_{\text{un}}(g,p(\mathbf{A}^{\text{aug}}),\mathbf{X}) + \alpha \mathcal{L}_{\text{fair}}(g,p(\mathbf{A}^{\text{aug}}),\mathbf{X}) \\
\textrm{subject to: } & f=\mbox{arg} \min\limits_{f} \mathcal{L}_{\text{fair}}(g,f(\mathbf{A}),\mathbf{X})\quad \mathbf{A}^{\text{aug}}=f(\mathbf{A})\\
& ||\mathbf{A}-\mathbf{A^{*}}||_{0}\leq  N.
\end{aligned}
\end{equation}


\subsection{Fair Edge Augmentation}

As shown in Section~\ref{sec:reason}, edge removal can increase the homophily ratio, thereby affecting representation fairness among local neighbors. To address this, $f$ targets on adding \textit{inter-group links} within local neighborhoods in $\mathcal{G}_f$ to mitigate biases introduced by edge removal. Specifically, We introduce a Boolean perturbation matrix $\mathbf{M}^\text{aug} \in \{0, 1\}$
to encode whether or not an edge in $\graph$ is modified. That is, the
edge connecting nodes $i$ and $j$ is added,
if and only if $\mathbf{M}^\text{aug}_{ij} = \mathbf{M}^\text{aug}_{ji} = 1$.  Given the adjacency matrix $\mathbf{A}$, this process can be described as :
\begin{equation}
\label{eq: perturb}
\begin{aligned}
\mathbf{A}^\text{aug}&=\mathbf{A}+(\mathbf{J}-\mathbf{I}-\mathbf{A}) \circ \mathbf{M}^\text{aug}.
\end{aligned}
\end{equation}

Here $\mathbf{J}$ denotes an all-one matrix. We exclude all edges in $\mathcal{E}_f$ from $\mathbf{A}$ and set the corresponding entries in $\mathbf{M}^\text{aug}$ to zero to prevent reintroduction. Due to the discrete nature of $\mathbf{M}$, we relax edge weights from binary variables to continuous variables
in the range $(0,1)$ and adopt the reparameterization trick for gradient-based optimization. Specifically, we sample $\mathbf{M}^\text{aug} \sim \text{Bernoulli}(\tilde{\mathbf{M}}^\text{aug})$, where $\tilde{\mathbf{M}}^\text{aug}$ represents the predicted probabilities of edge additions. For each pair of nodes $(i,j)$, the embeddings $\vz_i$ and $\vz_j$ of $\ori$ are used to estimate the probability of the edge $e_{ij}$. After the $\mathbf{M}^\text{aug}$ is sampled, only the \textit{inter-group links} are incorporated. To support the end-to-end training, we leverage the Gumbel-Softmax trick\cite{huijben2022review} to approximate the non-differentiable Bernoulli sampling:

\begin{equation}
\label{eq: aug}
\begin{aligned}
 \widetilde{\mathbf{M}}^{\text{aug}}&=\sigma\left(\frac{f\bigl([\vz_i;\vz_j]\bigr)+f\bigl([\vz_j;\vz_i]\bigr)}{2}\right) \\
\mathbf{M}^{\text{aug}}&=\frac{1}{1+\exp\Bigl(-(\log(\widetilde{\mathbf{M}}^{\text{aug}})+\mathbf{G})/\tau\Bigr)}
\end{aligned}
\end{equation}
where $f$ denotes a multi-layer perceptron(MLP), $[;]$ indicates concatenation and $\sigma$ is the sigmoid function. We ensure that $\mathbf{M}=\mathbf{M}^{T}$ in Equation~\ref{eq: aug} to maintain the symmetry of the perturbation matrix. Given the predicted probabilities $\widetilde{\mathbf{M}}^{\text{aug}}$, the relaxed Bernoulli sampling yields a continuous approximation, where $\tau$ is a temperature hyperparameter and $\mathbf{G} \sim \text{Gumbel}(0,1)$ is sampled from the standard Gumbel distribution.


As our objective is to generate fair augmentations by adding  edges, the ideal augmenter $f$ targets on\textbf{ }finding the optimal structure $\mathbf{A}^{\text{aug}}=f(\mathbf{A})$ that achieves representation fairness. However, we cannot achieve it via supervised training because there is no ground truth indicating which edges lead to fair representation and should be added. To address this issue, we propose to use a contrastive loss to optimize the augmenter $f$.

Inspired by
~\cite{kose2022fair,wu2021federated,zhang2025credit}, we propose a contrastive loss which explicitly penalizes $\mathbf{A}^{\text{aug}}$ for increasing the edge probability between nodes sharing the same sensitive feature. For clarity, we treat each node $i$ as an anchor and define the following pairs based on its relationship with other samples. Specifically:

\begin{itemize}
\item \textit{Intra positive pairs:} refers to pairs of an anchor and its connected nodes that share the same sensitive features. $V_{intra}^+(i)=\{j: \mathbf{A}^{\text{aug}}[i,j]=1|S(i)=S(j)\}.$
\item \textit{Inter positive pairs:} refers to pairs of anchors and their connected nodes that share different sensitive features. $V_{inter}^+(i)=\{j:\mathbf{A}^{\text{aug}}[i,j]=1|S(i)\neq S(j)\}.$
\item \textit{Intra negative pairs:}  refers to pairs of an anchor and its non-connected nodes that share different sensitive features. $V_{intra}^-(i)=\{j:\mathbf{A}^{\text{aug}}[i,j]=0|S(i)=S(j)\}.$
\end{itemize}
For each anchor, our key idea is to ensure that positive and negative samples share the same sensitive attributes as the anchor, rendering sensitive features uninformative for link probability. We define the $V_{intra}^-(i)$ as negative pairs, while treating $V_{intra}^+(i)$ and $V_{inter}^+(i)$ as positive pairs. Based on this, we design $\mathcal{L}_{\text{fair}}$ to enhance the link probability between the anchor and nodes in positive pairs relative to negative pairs. It is formulated as follows:

\begin{equation}
\mathcal{L}_{\text{fair}}=\sum_{v_i \in \V} \frac{-1}{|V_P(i)|}\sum_{j\in V_P(i)}\mbox{log}\frac{\mbox{exp}(\vz^\text{aug}_j \cdot \vz^\text{aug}_i)}{\sum\limits_{k\in V_{intra}^-(i)} \mbox{exp}(\vz^\text{aug}_i \cdot \vz^\text{aug}_k)}
\end{equation}

Here $V_P(i)=V_{intra}^+(i)\cup V_{inter}^+(i)$ and $\vz^\text{aug}=\ori(\mathbf{X},\mathbf{A}^\text{aug})$ where $\vz^\text{aug}$ represents the node embedding in $\mathbf{A}^\text{aug}$. The $\ori$ is fixed during the optimization of $\mathcal{L}_{\text{fair}}$. 





\subsection{Sparse Structure Unlearning}
To achieve fair unlearning, we consider finding an optimal structure by eliminating redundant edges from $\mathbf{A}^\text{aug}$, while keeping the unbiased and informative ones. Following other approximate-based unlearning methods, we also adopt a learnable mechanism to adjust the original model for the target. Specifically, we learn a pruner over $\mathbf{A}^\text{aug}$ to achieve the fair unlearning target. We optimize for the graph adjacency as follows:


\begin{equation}
\small
\label{eq:sparse}
\begin{aligned}
\un, p &= \mbox{arg} \min\limits_{\un, p}\mathcal{L}_{\text{un}}(p(\mathbf{A}^\text{aug}),\mathbf{X}) + \alpha \mathcal{L}_{\text{fair}}(p(\mathbf{A}^\text{aug}),\mathbf{X})+\beta \mathcal{L}_{\text{dist}}. \\
\end{aligned}
\end{equation}
Here, we set $\mathcal{L}_{\text{dist}}=||p(\mathbf{A}^{\text{aug}})-\mathbf{A}||_{1}$, where $||\cdot||_1$ is the $L^1$-norm, to constrain the number of edge modifications from the original adjacency matrix $\mathbf{A}$. Similar to $f$, the pruner $p$ is an MLP to generate the mask matrix $\mathbf{M}^p$ according to Equation \ref{eq: aug}. 
\begin{equation}
\label{eq: prun}
\mathbf{M}^p = \frac{1}{1+\mbox{exp}(-(\mbox{log}\widetilde{\mathbf{M}}^{p}+\mathbf{G})/\tau)}, \quad\mathbf{A}^{*} = \mathbf{A}^{\text{aug}}\odot \mathbf{M}^p.
\end{equation}
Note that $\widetilde{\mathbf{M}}^{p}$ is constructed using embeddings on $\mathbf{A}^\text{aug}$ with $\vz^{\text{aug}}=g(\mathbf{X},\mathbf{A}^\text{aug})$. Building on this, our method could be seamlessly combined with any graph unlearning loss function as $\mathcal{L}_{\text{un}}$~\cite{cheng2023gnndelete}\cite{li2024towards}\cite{yang2023contrastive}. 

Here we adopt the $\mathcal{L}_{\text{un}}$ from GNNDelete~\cite{cheng2023gnndelete}, which formulates the unlearning loss into two properties. For each deleted edges $e_{ij}$:
\begin{itemize}
    \item \textit{Deleted Edge Consistency}, where deleted edges should have a similar predicted probability to randomly sampled unconnected edges.\\ 
$\mathcal{L}_{\text{DEC}} = \mathcal{L}_{\text{MSE}}(\{ [\vz^{p}_i; \vz^{p}_j] | e_{ij} \in \mathcal{E}_f \}, \{ [\vz^{\text{aug}}_i; \vz^{\text{aug}}_j] | i, j \in \mathcal{V}\})$. 
    \item \textit{Neighborhood Influence}, where node embeddings post-un\-learn\-ing should be similar to prior-unlearning.\\ $\mathcal{L}_{\text{NI}} = \sum \Big( \mathcal{L}_{\text{MSE}}(\vz^{p}_u, \vz^{\text{aug}}_u) | u \in \mathcal{S}_{ij} \Big)$, where $S_{ij}$ is the local enclosing subgraph of $e_{ij}$ and $\mathbf{Z}^{p}=\un(\mathbf{X},\mathbf{A}^{*})$.
 \end{itemize}   
Finally, $\mathcal{L}_{\text{un}}=\lambda\mathcal{L}_{\text{DEC}}+(1-\lambda)\mathcal{L}_{\text{NI}}$; following GNNDelete, we set the optimal $\lambda=0.5$.

We present a theoretical observation demonstrating how the sparsification operator can facilitate unlearning.
\begin{theorem}(Bounding edge prediction of unlearned model $\un$ by $\ori$) Let $e_{ij}$ be an edge to be removed, and $\mathbf{W}$ be the last layer weight matrix in $\ori$. Then the norm difference between the dot product of the node representations $\vz_i, \vz_j$ from $\ori$ and  $\vz'_i, \vz'_j$ from the unlearned model $\un$ is bounded by:
\begin{align}
\left<\vz_i, \vz_j\right>- \left<\vz_i', \vz_j'\right> 
&\leq(\frac{1}{2} \|\mathbf{W}_{D}^{L}\|^{2}-1)\|\vz_i-\vz_j\|^{2}+\|\mathbf{W}_{D}^{L}\mathbf{W}\|^{2}\|\Delta\|^{2}
\,
\end{align} where $\Delta = \sum_{k\in \mathcal{C}_i } \mathbf{h}_{k} ^{L-1}-\sum_{k\in \mathcal{C}_j } \mathbf{h}_{k} ^{L-1}$.
\end{theorem}
Here $\mathcal{C}_i$ and $\mathcal{C}_j$ represent the common neighbors with masked edges connecting to nodes $i$ and $j$, respectively. In GNNDelete, $\mathbf{W}_D^{L}$ denotes the deletion matrix at layer $L$, and $\mathbf{h}_{k}^{L-1}$ represents the embedding of node $k$ from the previous layer. Detailed derivations are shown in the Appendix. The first term in the bound ensures the stability of the deletion operator, while the second term suggests that masking edges from common neighbors can enlarge the gap between $\ori$ and $\un$ in predicted probability of $e_{ij}$, thereby enhancing the unlearning capability.

To sum up, the training process can be described as a Bi-level optimization(Equation~\ref{eq: Two-step Goal}), where $\alpha,\beta$ are hyperparameters. In each iteration, we first update $f$ to minimize $\mathcal{L}_{\text{fair}}$ while keeping $p$ and $g$ fixed, then update both $g$ and $p$. The training continues until convergence. An overview of \method is illustrated in Figure~\ref{fig:method}.

\subsection{\textsc{FROG-Joint}: Baseline}
Instead of the Bi-Level optimization, we also introduce a Joint optimization algorithm to solve Equation~\ref{eq:Main Goal} as a baseline. Following~\cite{jin2021graph}, we directly model the $\mathbf{A}^{*}$ as a function of the embeddings as Equation~\ref{eq: aug}. In this way, the number of parameters for modeling graph structure no longer depends on the number
of nodes, hence avoid learning $O(|\V|^2)$ parameters and renders the algorithm applicable to large-scale graphs.

\section{Worst-Case Evaluation}
\label{sec:evaluation}

Inspired by~\cite{fan2024challenging}, we evaluate unlearning methods with two different challenging settings: $(i)$ \textit{worst-case unlearning}, where $\mathcal{G}_f$ consists of edges that are hardest to forget, and $(ii)$ \textit{worst-case fairness}, where $\mathcal{G}_f$ consists of edges that negatively impacts fairness on $\mathcal{G}_r$ post-unlearning.
Let $w \in \{0, 1\}^{|\mathcal{E}|}$ denote a binary mask over all edges, where $w_{i,j} = 1$ indicates that the edge $e_{ij}$ belongs to the forget set.
Our objective is to optimize $w$ such that $\mathcal{G}_f$ contains all hard-to-forget edges or those critical for fairness.


\noindent \textbf{\textit{Worst-case unlearning.}} We select the forget set $\mathcal{G}_f$ to maximize the difficulty of effective unlearning. In other words, after unlearning $\mathcal{G}_f$, the unlearned model will exhibit a low loss on $\mathcal{G}_f$, indicating a failure to fully eliminate the influence of $\mathcal{G}_f$ from the model. Specifically, we solve:
\begin{align}
\label{eq:worst-unlearn}
&\min_{w \in S} \sum_{e_{ij} \in \mathcal{G}} [w_{ij} \mathcal{L}_{\text{LP}}(\un; \vz_i,\vz_j)] + \gamma \|w\|_2^2
\\
&\text{subject to: } \quad \un = \arg \min_g \mathcal{L}_{\text{un}}(g; w),
\end{align} where $\mathcal{L}_{\text{LP}}$ is the link prediction loss.

In the upper-level optimization, we search for the edges defined by the binary edge mask $w$ that yield the worst unlearning performance. In other words, the loss on the forget set $\sum_{e_{ij}} \mathcal{G} \in [w_{ij} \mathcal{L}_{\text{LP}}(\un; \vz_i,\vz_j)]$ is minimized (unsuccessful unlearning). We additionally regularize the size of $w$ with the $L_2$ norm, since unlearning requests are much sparser than the original dataset. \emph{In the lower-level} optimization, the unlearned model $\un$ is obtained based on the forget set selected by $w$. 


\smallskip
\noindent \textbf{\textit{Worst-case fairness.}} We choose the forget set $\mathcal{G}_f$ to maximize fairness degradation. That is, after unlearning $\mathcal{G}_f$, the $\mathcal{L}_{\text{fair}}$ on the retained set $\mathcal{G}_r$ is maximized, indicating a failure to preserve the fairness. We solve:
\begin{align}
\label{eq:worst-fair}
&\max_{w \in S} \sum_{e_{ij} \in \mathcal{G}}[(1 - w_{ij}) \mathcal{L}_{\text{fair}}(\un; \vz_i, \vz_j)] + \gamma \|w\|_2^2
\\
&\text{subject to } \quad \un = \arg \min_g \mathcal{L}_{\text{un}}(g; w).
\end{align}

%% file: 070experiments.tex
\begin{figure*}[t]
    \centering
    \includegraphics[width=1\linewidth]{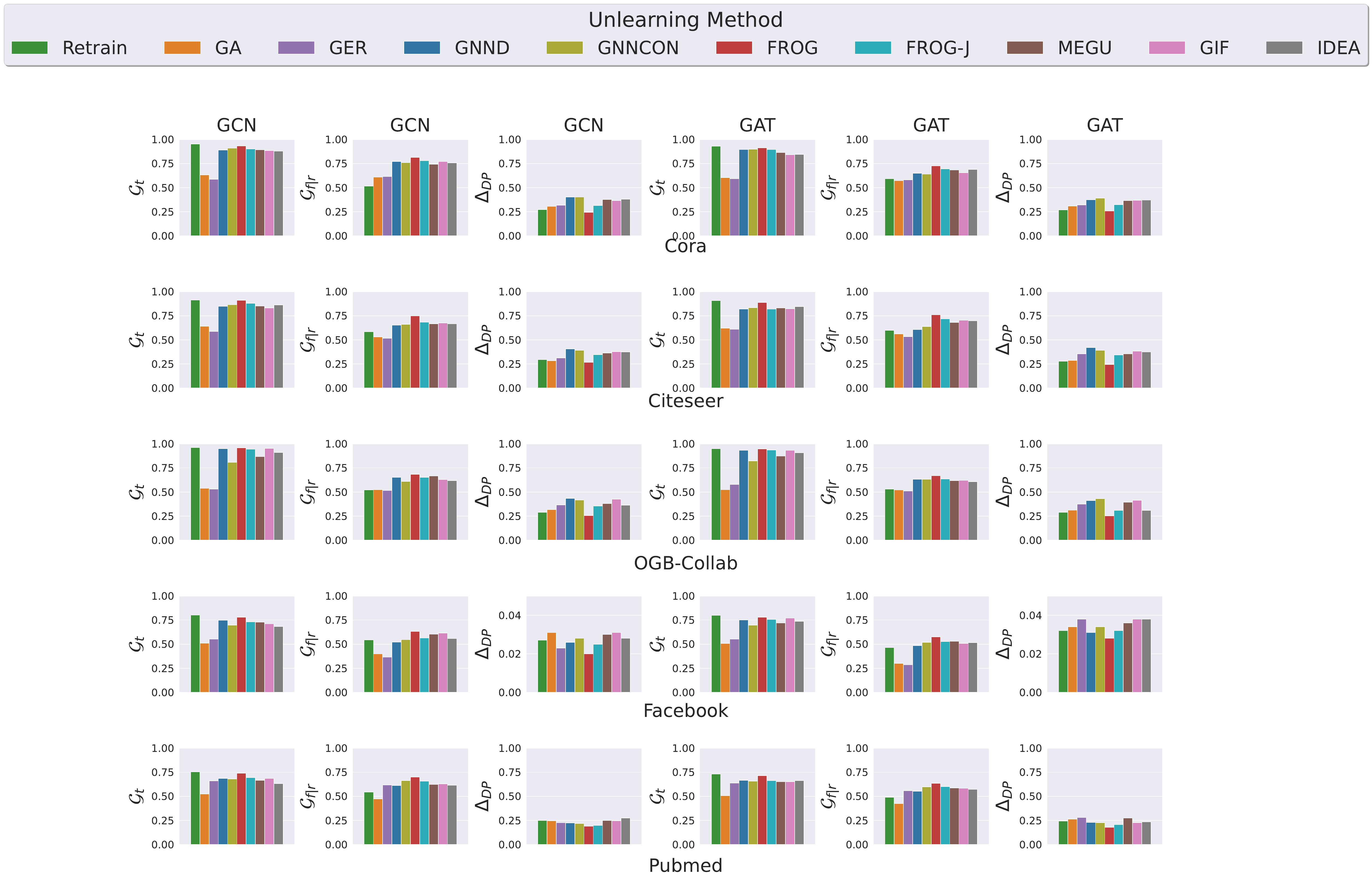}
    \caption{Effectiveness and fairness performance of edge unlearning on Cora, CiteSeer, OGB-Collab, Facebook, and Pubmed.}
    \label{fig:edge_classification}
\end{figure*}

\section{Experiments}
\label{sec:experiments}
To evaluate the effectiveness of our proposed model, we examine the following questions:
\begin{itemize}
\item \textbf{RQ1:} How is the unlearning efficacy of \method and its impact on graph fairness under uniform cases?
\item \textbf{RQ2:} How is the unlearning efficacy of \method and its impact on graph fairness under worst-case scenarios?

\end{itemize}

\noindent\textbf{\textit{Datasets.}} We perform experiments over the following real-world datasets: CiteSeer~\citep{bojchevski2018deep}, Cora~\citep{bojchevski2018deep}, OGB-Collab~\cite{cheng2023gnndelete}, Facebook$\#1684$. ~\cite{li2021dyadic} and Pubmed~\cite{cheng2023gnndelete}. 
Facebook$\#1684$ is a social ego network from the Facebook app, and we select gender as the sensitive feature. The rest citation networks, each vertex represents an article with descriptions as features. A link stands for a citation. We set the category of an article as the sensitive attribute. 
These datasets usually serve as a commonly used
benchmark datasets for GNN performance over link prediction and node classification tasks.

\smallskip
\noindent \textbf{\textit{Baselines.}} We compare \method to the following baselines: $(i)$ Retrain, which refers to training from scratch; $(ii)$ GA~\citep{Golatkar_2020_CVPR}, which performs gradient ascent on $\mathcal{G}_f$; $(iii)$ GER~\citep{chen2022graph}, a re-training-based machine unlearning
method for graphs; $(iv)$ GNNDelete~\citep{cheng2023gnndelete}, an approximate graph unlearning method that treats deleted edges as unconnected node pairs; $(v)$ GNNCON~\citep{yang2023contrastive} a contrastive learning based method. Finally, $(vi)$ MEGU~\cite{li2024towards}, $(vii)$ GIF~\cite{wu2023gif}, and $(viii)$ IDEA~\cite{dong2024idea}, which represent advanced graph unlearning methods.

\smallskip
\noindent \textbf{\textit{Unlearning Task.}} We evaluate \method under two unlearning tasks: $(i)$ \textit{node unlearning}, where a subset of nodes $\mathcal{N}_f \in \mathcal{N}$ and all their associated edges are unlearned from $g_w$; and $(ii)$ \textit{edge unlearning}, where a subset of edges $\mathcal{E}_f \in \mathcal{E}$ are unlearned from $g_w$. In line with prior works, the forget set comprises 5\% of the entire dataset.

\smallskip
\noindent \textbf{\textit{Sampling of Forget Set.}}
For worst-case unlearning, the forget set $\mathcal{G}_f$ is chosen through optimization according to \ref{eq:worst-unlearn} and \ref{eq:worst-fair}. For uniform cases, the forget set $\mathcal{G}_f$ is randomly selected within 3-hops of $\mathcal{G}_t$. Due to the limited space, we only conduct edge unlearning in the worst-case evaluation.

\smallskip
\noindent \textbf{\textit{Evaluation Metrics.}} We evaluate the unlearned model's performance from the following two perspectives:
\begin{itemize}
    \item \textit{Effectiveness-oriented}, which probes if $\mathcal{G}_f$ is unlearned from $\mathcal{G}_w$ while preserving model utility. Specifically, we compute $(i)$ the test set AUROC $\mathcal{G}_t (\uparrow)$, $(ii)$ the forget-retain knowledge gap $\mathcal{G}_{f|r} (\uparrow)$~\citep{cheng2023gnndelete,cheng2024mubench} which quantifies how well a model distinguishes unlearned and retained data. Specifically, the knowledge gap is computed as the AUROC score with the prediction logits of $\mathcal{G}_r$ and $\mathcal{G}_f$, and their labels as 1 and 0, respectively. We also demonstrate the success rate of membership inference attack (MIA) by $\mbox{MI}(\uparrow)$.
    

    \item \textit{Fairness-oriented}, which evaluates the fairness of the unlearned model. In node classification, we focus on Equalized Odds (EO) $\Delta_{EO}=|p(\hat{y}=1|y=1,s=1)-p(\hat{y}=1|y=1,s=0)|$ and Statistical Parity (SP) $\Delta_{SP}=|p(\hat{y}=1|s=1)-p(\hat{y}=1|s=0)|$~\cite{spinelli2021fairdrop} to evaluate the group disparity. In the link prediction scenario, we directly use Demographic Parity $\Delta_{DP}$ to measure the dyadic fairness. 
\end{itemize}

%% file: 080results.tex
\section{Results}

\begin{figure}[t]
    \centering
    \includegraphics[width=0.9\linewidth]{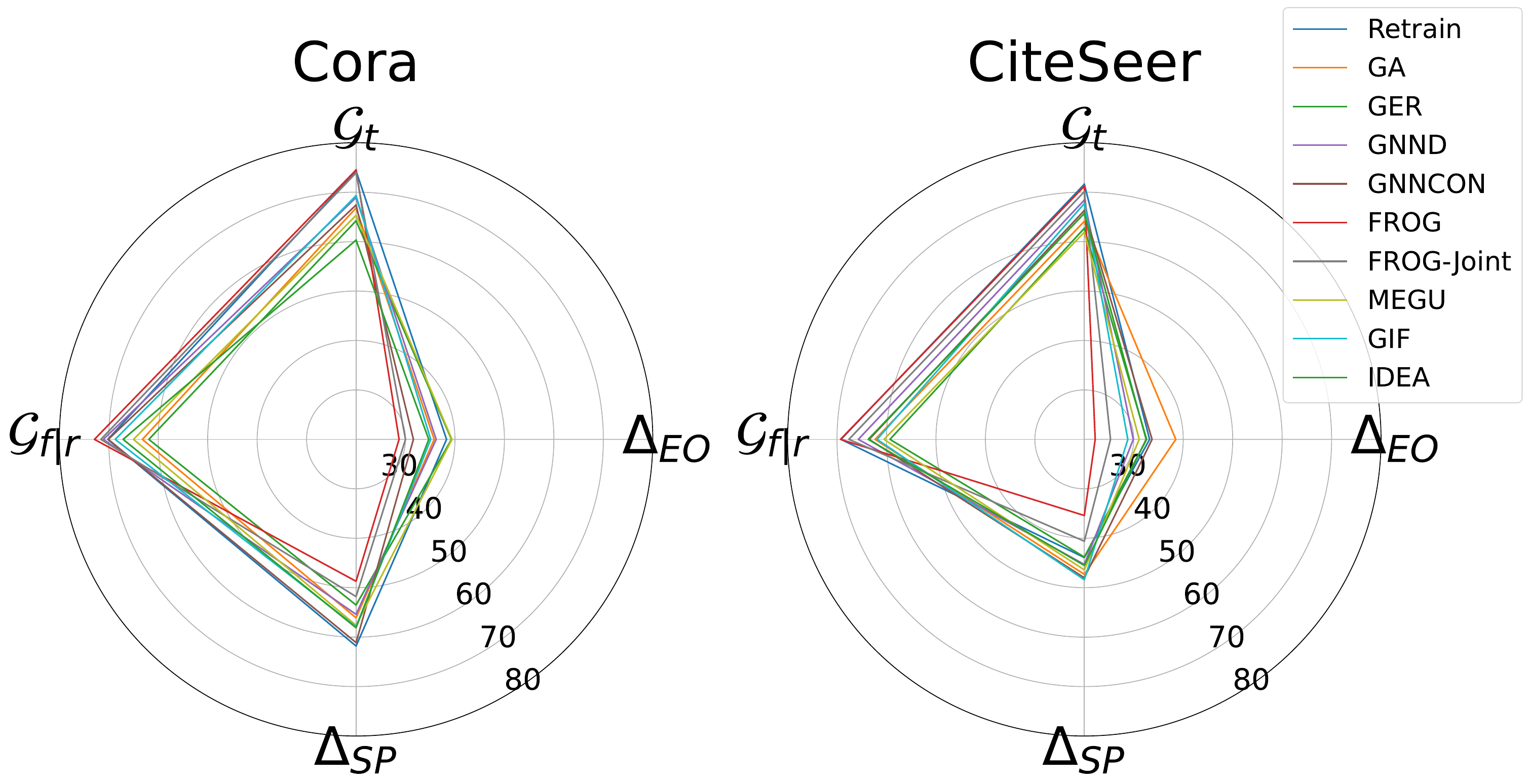}
    \caption{Effectiveness and fairness performance of node unlearning.}
    \label{fig:nodeunlearning}
\end{figure}

\subsection{RQ1: FROG Performance Under Uniform Removal}

\noindent \textbf{\textit{Existing graph unlearning methods hurt fairness.}} Existing graph unlearning methods, though effective, detrimentally impact graph fairness post-unlearning. Results in Figure~\ref{fig:edge_classification} show that GA, GNND, GNNCON have DP degradation of $-0.089$, $-0.16$, $-0.21$, $-0.20$, respectively. Notably, even Retrain compromises fairness by $-0.064$. GNND and GNNCON, though effective in unlearning with competitive scores on $\mathcal{G}_t$ and $\mathcal{G}_{f|r}$, suffer from the most significant degradation of fairness. This highlights that existing state-of-the-art graph unlearning models have overlooked graph fairness as an important factor to consider, which may hinder their application in fairness-concerned scenarios.

\smallskip
\noindent \textbf{\textit{\method is effective in unlearning.}}
In edge unlearning, \method successfully distinguishes unlearned edges from retained edges measured by $\mathcal{G}_{f|r}$. As shown in Figure~\ref{fig:edge_classification}, when randomly removing 5\% edges, \method outperforms Retrain, GA, GER, GNND, and GNNCON by $10.0$, $17.2$, $19.7$, $8.6$, $5.9$ absolute points, respectively. Under the node unlearning setting, as shown in Figure~\ref{fig:nodeunlearning}, \method outperforms Retrain, GA, GER, GNND, and GNNCON by $5.7$, $9.7$, $13.0$, $5.8$, $3.9$ absolute points, respectively, when deleting 5\% of nodes. These results indicate that \method exhibits more successful targeted knowledge removal of $\mathcal{G}_f$ than baselines. Meanwhile, \method preserves model utility on downstream prediction tasks measured by $\mathcal{G}_t$, outperforming GA, GER, GNND, IDEA, MEGU by $32.2$, $31.0$, $6.0$, $4.8$,$4.2$,$4.9$ absolute points respectively, when deleting 5\% of edges (Figure~\ref{fig:edge_classification}). \method is even comparable to Retrain with a trivial gap of $2.2$.

\begin{figure}[t]
    \centering
    \includegraphics[width=1\linewidth] 
     {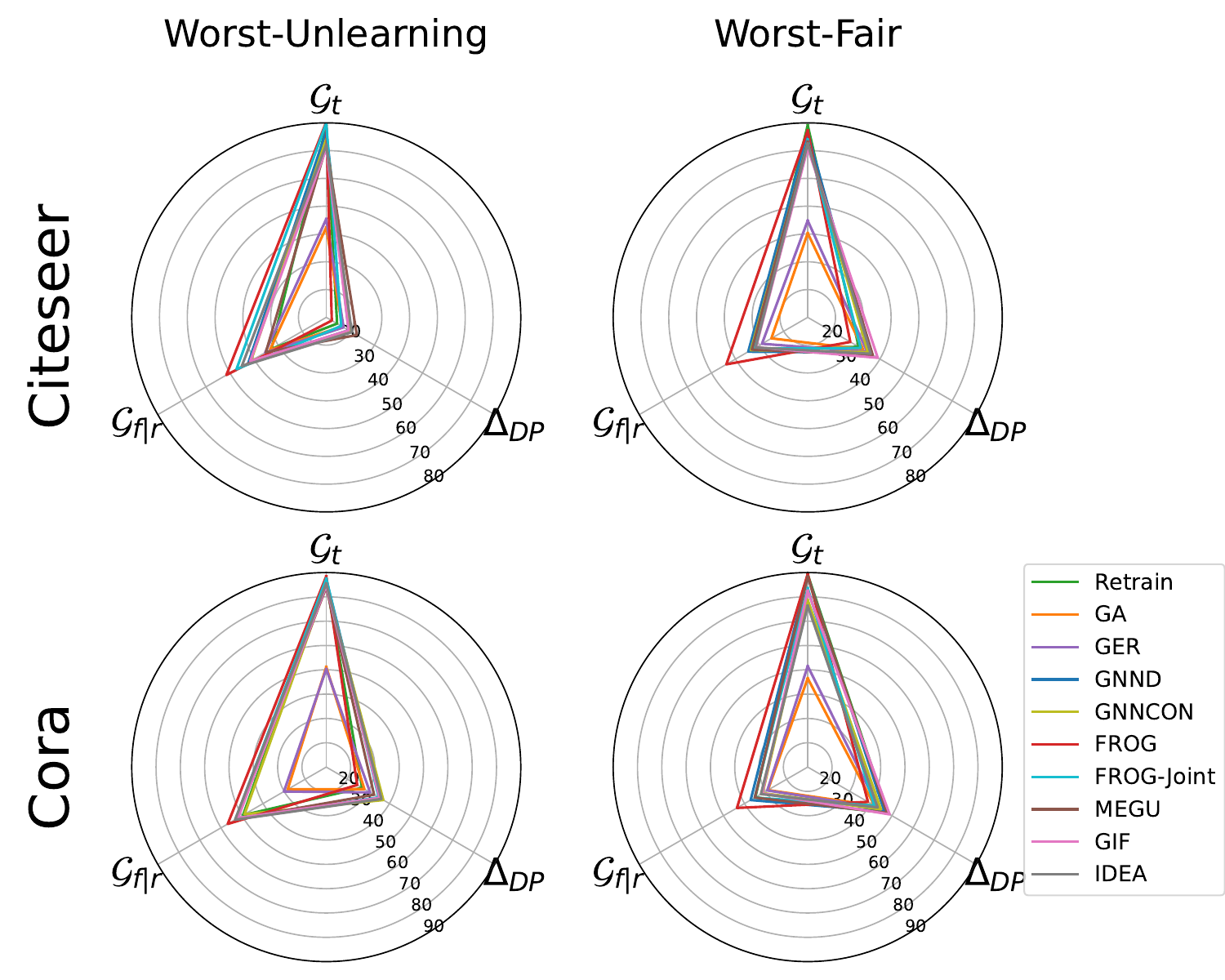}
    \caption{Worst-case unlearning (left) and worst-case fairness (right) performance on CiteSeer and Cora.}
    \label{fig:worstcase}
\end{figure}

\smallskip
\noindent \textbf{\textit{\method preserves fairness during unlearning.}}
\method consistently achieves the best fairness performance in terms of $\Delta_{EO}$, $\Delta_{SP}$, and $\Delta_{DP}$ on both edge and node unlearning tasks. For instance, in the edge unlearning task on the Cora dataset, \method reduces $\Delta_{DP}$ by $21.6\%$, $39.5\%$, $35.4\%$, and $36.9\%$ compared to GA, GNND, MEDU, and IDEA, respectively (Figure~\ref{fig:edge_classification}). In the node unlearning task, compared with IDEA, \method reduces $\Delta_{EO}$ by $23.9\%$ on Cora and $46.3\%$ on CiteSeer, with comparable effectiveness (Figure~\ref{fig:nodeunlearning}).
We attribute this superior performance over baselines to the fairness-aware design of the proposed method. Specifically, the optimization-based graph structure modification aims to find the optimal topology that results in both successful unlearning and minimum damage to fairness.

\begin{figure}[t]
    \centering
    \includegraphics[width=1\linewidth]{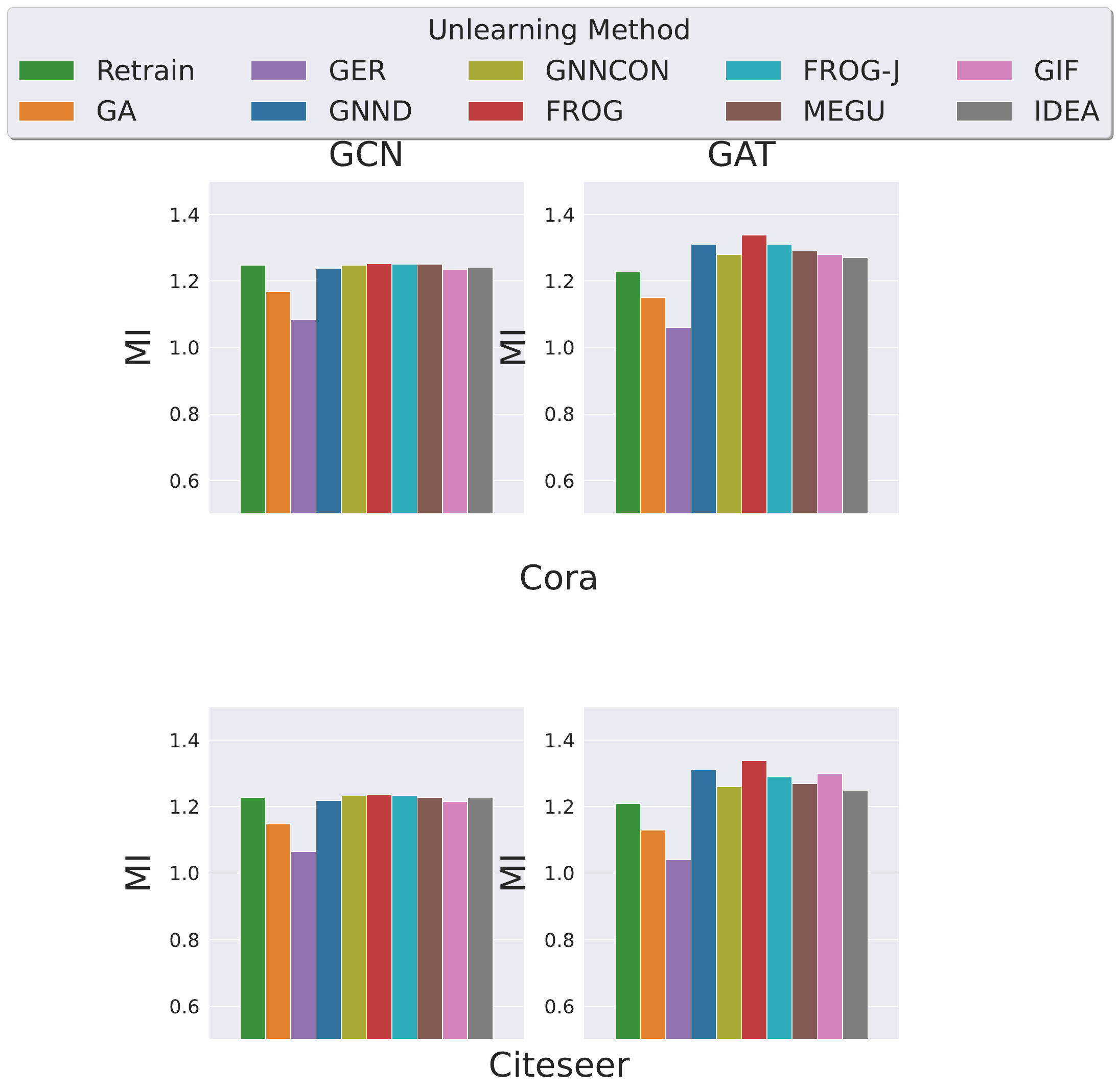}
    \caption{Reported is the MI attack ratio. The left and right columns use GCN and GAT as backbones, respectively. }
    \label{fig:MI}
\end{figure}

\smallskip
\noindent \textbf{\textit{\method effectively hides deleted information.}} Following~\cite{cheng2023gnndelete}, we delete 100 nodes and their associated edges from the training data and adopt Membership Inference attacks~\cite{liu2024please} on two datasets. Reported is the MI attack ratio. As shown in Figure~\ref{fig:MI}, \method outperforms other baseline methods, highlighting its effectiveness in hiding deleted information. Across two GNN architectures, we find that \method achieves the highest $\mbox{MI}$ ratio score of all baselines. Specifically, it outperforms GA, GER, GNNDelete, GIE, and IDEA by $0.089$, $0.162$, $0.035$, $0.134$, $0.086$.

\smallskip
\noindent \textbf{\textit{Time Efficiency.}} We demonstrate that \method is time-efficient compared to most unlearning baselines, as shown in Table~\ref{tab:time}. Specifically, \method is $\mathbf{11.5}\times$ faster than Retrain on CiteSeer and $\mathbf{1.25}\times$ faster than IDEA and GNNCON. Although slightly slower than GNNDelete due to its structure learning component, \method achieves significantly better predictive performance and fairness.

\smallskip
\noindent \textbf{\textit{Ablation Study.}} We examine the effect of the hyperparameter $\alpha$, which balances the unlearning and fairness objectives, as shown in Table~\ref{tab:ablation}. The results highlight the necessity of both $\mathcal{L}_{\text{un}}$ and $\mathcal{L}_{\text{fair}}$ for achieving a favorable AUROC–$\Delta_{DP}$ trade-off. As $\alpha$ increases, \method places greater emphasis on reducing $\Delta_{DP}$. When $\alpha \leq 0.2$, edge unlearning performance remains stable; however, for $\alpha > 0.4$, AUROC drops sharply. To balance the trade-off, we set $\alpha=0.2$ in our experiments.

\begin{table}
\small
\setlength{\tabcolsep}{4pt}
    \centering
    \caption{Time spent when unlearning 100 edges on CiteSeer.}
    \begin{tabular}{l|c|c|c|c|c}
    \toprule
    Method &  Retrain &    GA &   GER &  GNNDelete &  GNNCON \\
    \midrule
    Time (hr) &  0.63 &  0.31 &  0.26 & 0.025 &  0.05   \\
    \midrule
    \midrule
    Method & MEGU &  GIF &  IDEA &  FROG &  FROG-Joint \\
    \midrule
    Time (hr) &  0.04 &  0.10 &  0.06 &  0.04 &    0.03 \\
    \bottomrule
    \end{tabular}
    \label{tab:time}
\end{table}

\subsection{RQ2: FROG Performance Under Worst-Case Scenarios}


\noindent{\textbf{\textit{Unlearning Performance.}}}
When $\mathcal{G}_f$ involves the set of edges that are the hardest to unlearn, we find that \method is more effective to differentiate between $\mathcal{G}_f$ and $\mathcal{G}_r$ than baselines. This shows that in worst case, \method still demonstrates more successful knowledge removal than existing graph unlearning methods, see Figure~\ref{fig:worstcase}. We attribute this to the graph sparsification process, which simplifies hard-to-forget graph parts~\citep{liu2024model,tan2024unlink,sun2023all}.

\smallskip
\noindent \textbf{\textit{Fairness Performance.}}
Similarly when $\mathcal{G}_f$ involves the set of edges that are introduces the largest fairness degradation post-unlearning, \method provides fairer representations than baselines, see Figure~\ref{fig:worstcase}. The advantage on fairness inherits from the edge addition process, which injects new heterogeneous edges and dramatically mitigates network segregation. These results highlights the robustness of \method to adversarial unlearning sets under extreme cases, making users more confident to apply \method in real-world applications.



\begin{table}[t]
\small
\centering
\caption{Ablation study of effect on $\alpha$ on CiteSeer. Best performance is \textbf{bold} and second best is \underline{underlined}.}
\begin{tabular}{l|ccc|ccc}
\toprule
\multirow{2}{*}{$\alpha$} & \multicolumn{3}{c|}{$|\mathcal{G}_f| = 2.5\%$} & \multicolumn{3}{c}{$|\mathcal{G}_f| = 5\%$} \\
\cmidrule{2-7}
 & $\mathcal{G}_t (\uparrow)$ & $\mathcal{G}_{f|r} (\uparrow)$ & $\Delta_{DP}$ $(\downarrow)$ & $\mathcal{G}_t (\uparrow)$ & $\mathcal{G}_{f|r} (\uparrow)$ & $\Delta_{DP}$ $(\downarrow)$ \\
\midrule
0.0 & \bf{90.7} & 79.0 & 32.9 & \bf{90.1} & \bf{71.3} & 37.5 \\
0.1 & \bf{90.7} & \underline{79.1} & 29.5 & 89.9 & 70.8 & 29.2 \\
0.2 & \underline{90.6} & \bf{79.6} & 25.8 & \bf{90.1} & \underline{71.0} & \underline{25.5} \\
0.4 & 89.1 & 76.2 & 23.7 & 89.9 & 70.4& \underline{25.5} \\
0.6 & 87.2 & 75.7 & \bf{23.5} & 89.2 & 70.1 & 25.7 \\
0.8 & 90.3 & 72.1 & \underline{26.7} & 88.9 & 69.5 & \bf{24.9} \\
\bottomrule
\end{tabular}
\label{tab:ablation}
\end{table}

\begin{figure}[t]
    \centering
    \includegraphics[width=1\linewidth]{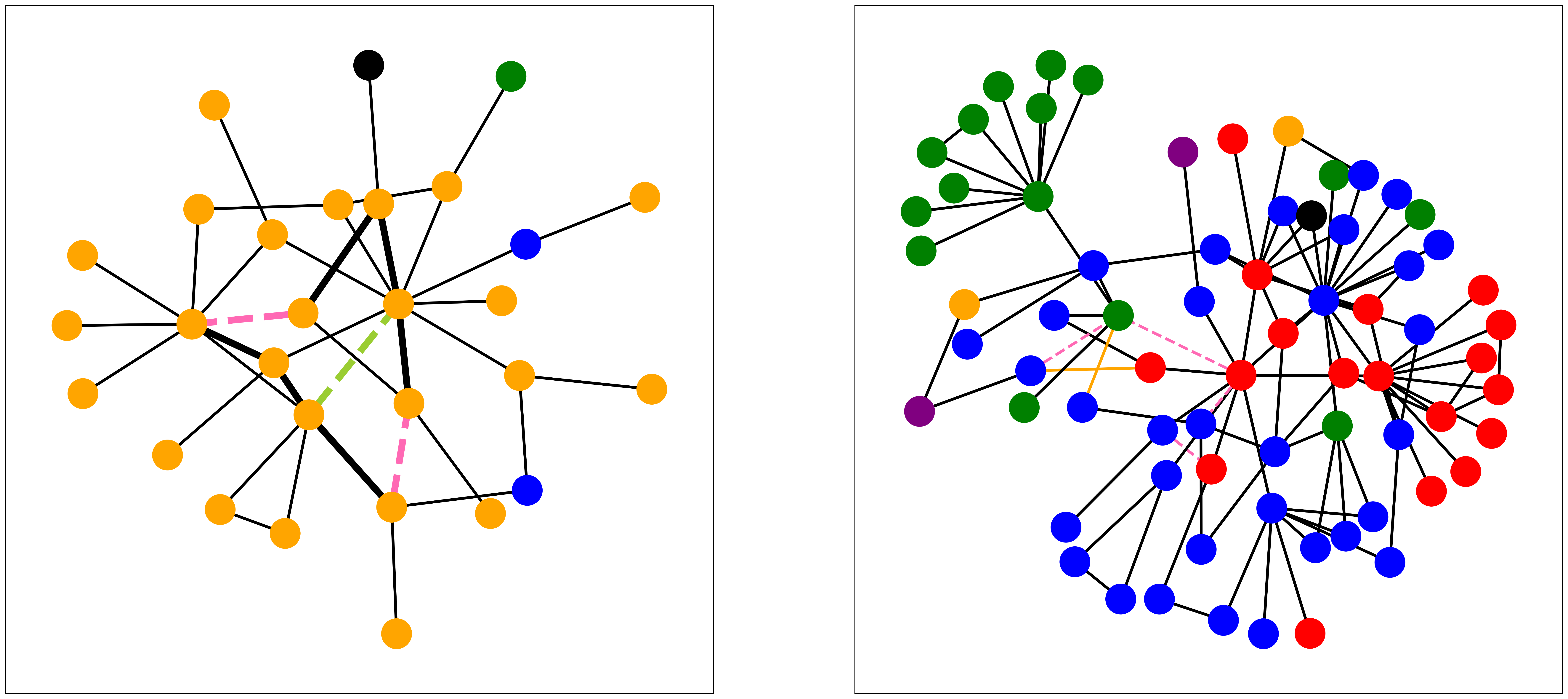}
    \caption{Illustration of our method under worst-case evaluation. \textbf{Pink dashed edges}: edges to be removed. \textbf{Green dashed edges}: edges masked by FROG. \textbf{Orange solid edges}: edges to be added by FROG.}
    \label{fig:worst}
\end{figure}

\smallskip
\noindent \textbf{\textit{Case Study.}} We present a case study in Figure~\ref{fig:worst}. 
As a practical example for edge unlearning, we evaluate the performance of our algorithm in worst-case scenarios, as shown in Figure~\ref{fig:worst}. 
In the left panel, the dashed pink edges represent hard-to-forget edges, identified using Equation~\ref{eq:worst-unlearn}. We observed that the two forgotten edges belong to loops (highlighted by bold edges) that continue facilitating message passing through their common neighbors even after removal. Consequently, these loops impede the unlearning of the target edges. To address this, our method proposes masking the dashed green edge, effectively breaking both loops simultaneously. In the right panel, the two worst-fairness edges obstruct message passing between different sensitive groups—one cluster dominated by green nodes and the other predominantly by blue and red nodes. To address this, the algorithm suggests adding two edges that connect the clusters without creating new loops.

%% file: 020related.tex
\section{Related Work}
\label{sec:related}


\noindent \textbf{\textit{Graph Unlearning.}}
Machine unlearning on graphs~\citep{chen2022graph} focuses on removing data influence from models. GraphEraser~\citep{chen2022graph} approaches graph unlearning by dividing graphs into multiple shards and retrain a separate GNN model on each shard. However, this can be inefficient on large graphs and dramatically hurt link prediction performances. 
UtU formulates graph unlearning as removing redundant edges~\citep{tan2024unlink}. \cite{cheng2024multidelete} develop a method to unlearn associations from multimodal graph-text data. CEU uses influence function for GNNs to achieve certified edge unlearning~\cite{ceu,huang2025prompt}.
However, how graph unlearning impacts the representation fairness of the retain graph remains unexplored. 

\smallskip
\noindent \textbf{\textit{Graph Fairness.}}
Numerous studies have explored fairness issues in graph learning. Fairwalk~\cite{rahman2019fairwalk} introduced a random walk-based graph embedding method that adjusts transition probabilities based on nodes' sensitive attributes. Then in ~\cite{liao2020graph}, they propose to use adversarial training on node embeddings to minimize the disparate parity. Then in~\cite{li2021dyadic}, the focus shifted to dyadic fairness in link prediction, emphasizing that predictive relationships between instances should remain independent of sensitive attributes.  Other works include fair collaborative filtering~\cite{yao2017beyond} in bipartite graphs and item recommendation tasks~\cite{chakraborty2019equality}.
Only a few works examine the interplay of unlearning and fairness or bias at the same time~\citep{chen2024debiasing,cheng2024mubench}.

\smallskip
\noindent \textbf{\textit{Adversarial Unlearning.}}
Several works try to stress-test unlearning methods under adversarial settings. On the method side, \cite{10.1145/3477495.3531820} investigates how adversarial training can help forget protected user attributes, such as demographic information. MUter~\citep{Liu_2023_ICCV} analyzes unlearning on adversarially trained models. 
On the evaluation side, \cite{goel2022towards} argues that unlearning should remove the generalization capability in addition to the data samples themselves. \cite{fan2024challenging} studies adversarial unlearned data and proposes an approach to find challenging forget samples through bi-level optimization. \cite{cheng2025tool} extends existing membership inference attacks to diverse data samples, even samples not in the training set, to check if a model has truly forgotten some knowledge. \cite{chen2024debiasing} uses counterfactual explanations to debias the machine unlearning algorithm in the classification task.


%% file: 090conclusion.tex
\section{Conclusion}
\label{sec:conclusion}

We presented \method as the first graph unlearning method that not only effectively removes graph elements but also preserves the fairness of the retained graph. 
We formulated this problem as a bi-level optimization task that jointly optimizes the unlearned model and the underlying graph topology. Experiments on four datasets demonstrated that \method can successfully unlearn information while preserving fairness. Furthermore, adversarial evaluations under challenging scenarios showed that \method outperforms existing methods, achieving robust performance.


%% file: 091appendix.tex



\input{092proof}

%% file: 092proof.tex
\section{Appendix}

\subsection{Proofs}
\textbf{Proof of Theorem 3.1:} 
For any pair of nodes coming from two different sensitive groups $v_i\in S_0, v_j\in S_1$, we have:
\begin{equation}
z_i-z_j=
\mathbf{W}\Bigg(\frac{1}{d_i+1}\sum_{v_p\in \mathcal{N}_i\cup v_i}X_p-\frac{1}{d_j+1}\sum_{v_q\in \mathcal{N}_j\cup v_j}X_q\Bigg).
\end{equation}
Based on the definition of $\rho$, among $|\mathcal{N}_i\cup v_i|=d_i+1$ neighboring nodes of $v_i$, $\rho(d_i+1)$ of them come from the same sensitive feature distribution as $v_i$ while $(1-\rho)(d_i+1)$ of them come from the opposite feature distribution as $v_j$, then we have:
 \begin{equation}
\begin{aligned}
\frac{1}{d_i+1}\sum_{v_p\in \mathcal{N}_i\cup v_i}X_p & \sim \mathcal{N}(\rho\mu^{0}+(1-\rho)\mu^{1},\frac{1}{d_i+1}(\rho \Sigma^{0}+(1-\rho)\Sigma^{1}));
\\
\frac{1}{d_j+1}\sum_{v_q\in \mathcal{N}_j\cup v_j}X_q & \sim \mathcal{N}(\rho\mu^{1}+(1-\rho)\mu^{0},\frac{1}{d_j+1}(\rho \Sigma^{1}+(1-\rho)\Sigma^{0})).
\end{aligned}
\end{equation}
 Consider $\vz_i-\vz_j$ follows a normal distribution, i.e.,  $\vz_i - \vz_j \sim \mathcal{N}(\mu,\Sigma)$, where 
 $\mu=(2\rho-1)\mathbf{W}(\mu^{0}-\mu^{1})=(2\rho-1)\mathbf{W}\delta$.
Let us denote $E_{i\in S_0}[z_i]=p$ and $E_{j\in S_1}[z_j]=q$. Thus:
\begin{equation}
\begin{aligned}
\Delta_{DP}
&=|E_{\substack{(i,j)\\ S_{i}= S_{j}}}[\vz_j\cdot \vz_i]-E_{\substack{(i,j)\\ S_{i}\neq S_{j}}}[\vz_i \cdot \vz_{j}]|\\
&= |p^{T}q-\Big(\frac{|S_0|^2}{|S_0|^2+|S_1|^2}p^{T}p+\frac{|S_1|^2}{|S_0|^2+|S_1|^2}\Big)q^{T}q|\\
&\leq E||q-p||_2\Big(\frac{|S_0|^2}{|S_0|^2+|S_1|^2}p+\frac{|S_1|^2}{|S_0|^2+|S_1|^2}q\Big).
\end{aligned}
\end{equation}
Since $E|p-q|=\dot(2\rho-1)\mathbf{W}\delta$, therefore it holds the following:
\[\Delta_{DP} \leq |K\cdot(2\rho-1)\mathbf{W}\delta|.\]





\noindent \textbf{Proof of Theorem 5.1:} 
We first consider the case without structural modifications. In GNNDelete, $\mathbf{W}_D^{L}$ denotes the deletion matrix at layer $L$, and $\mathbf{h}_{j}^{L-1}$ represents the embedding of node $j$ from the previous layer.
Following~\cite{cheng2023gnndelete}, We have $\vz_i = \sigma\Big(\sum_{j\in i \cup \mathcal{N}_i} \mathbf{W}\mathbf{h}_{j} ^{L-1}\Big),$ After normalization, we have:
\begin{multline}\label{eq:dot_product}
    \left<\vz_i, \vz_j\right>- \left<\vz_i', \vz_j'\right> 
= \frac{1}{2}\|\vz_i'-\vz_j'\|^{2} - \frac{1}{2}\|\vz_i-\vz_j\|^{2}.
\end{multline}
Considering the deletion matrix and~\ref{eq:dot_product}, we have:
\begin{align}\label{eq:bound_dot_product}
\|\vz_i'-\vz_j'\| &= \| \sigma(\mathbf{W}_{D}^{L}\vz_i) - \sigma(\mathbf{W}_{D}^{L}\vz_j) \| \\
     & \overset{\text{Lipschitz } \sigma} \leq \|\mathbf{W}_{D}^{L}\vz_i - \mathbf{W}_{D}^{L}\vz_j \| \\
     & \overset{\text{Cauchy-Schwartz}} \leq \|\mathbf{W}_{D}^{L}\| \| \vz_i - \vz_j \|
\end{align}
and applying that to Equation~\ref{eq:dot_product}:
\begin{align}\label{eq:dot_product_contd}
     \begin{split}
    \left<\vz_i, \vz_j\right>- \left<\vz_i', \vz_j'\right> \leq \frac{1}{2} ( \|\mathbf{W}_{D}^{L}\|^{2} - 1 ) \| \vz_i - \vz_j \|^{2}\\
   \end{split}.
\end{align}

Now we consider the structural modification. To forget edge $(i,j)$, let $\mathcal{C}$ be their common neighbors. $\mathcal{C}_i \in \mathcal{C}$ and $\mathcal{C}_j \in \mathcal{C}$ are nodes with masked links to $i$ and $j$, and $\mathcal{C}_{i \cap j}=\mathcal{C}_i\cap \mathcal{C}_j$. We can rewrite $z_i$ as: 
\[\vz_i = \mathbf{W}\Big(\sum_{u\in \mathcal{C}_i } \mathbf{h}_{u} ^{L-1}+\sum_{v\in \mathcal{C}_j } \mathbf{h}_{v} ^{L-1}+\sum_{k\in \mathcal{C}_{i\bigcap j} } \mathbf{h}_{k} ^{L-1}+O_i\Big), 
\]
\[\vz_j = \mathbf{W}\Big(\sum_{u\in \mathcal{C}_i } \mathbf{h}_{u} ^{L-1}+\sum_{v\in \mathcal{C}_j } \mathbf{h}_{j} ^{L-1}+\sum_{k\in \mathcal{C}_{i\bigcap j} } \mathbf{h}_{k} ^{L-1}+O_j\Big), 
\]
Here, $O_i$ denotes the sum of representations of nodes that are neighbors of $i$ but not linked to $j$. After masking the redundant edges, we have:
\[\vz'_i = \sigma\Big(\mathbf{W}_{D}\mathbf{W}(\sum_{v\in \mathcal{C}_j } \mathbf{h}_{j} ^{L-1}+O_i)\Big), \quad \vz'_j = \sigma\Big(\mathbf{W}_{D}\mathbf{W}(\sum_{u\in \mathcal{C}_i } \mathbf{h}_{u} ^{L-1}+O_j)\Big)
\]
\begin{align}
\label{eq:bound_dot_product}
\|\vz_i'-\vz_j'\| 
&=\|\sigma\Big(\mathbf{W}_{D}\mathbf{W}(\sum_{v\in \mathcal{C}_j } \mathbf{h}_{j} ^{L-1}+O_i)\Big)- \sigma\Big(\mathbf{W}_{D}\mathbf{W}(\sum_{u\in \mathcal{C}_i } \mathbf{h}_{u} ^{L-1}+O_j)\Big)\|\\
&\leq\|\sigma\Big(\mathbf{W}_{D}^{L}\mathbf{W}(O_i-O_j+\sum_{v\in \mathcal{C}_j } \mathbf{h}_{v} ^{L-1}-\sum_{u\in \mathcal{C}_i } \mathbf{h}_{u} ^{L-1})\Big)\|\\
&=\|\sigma\Big(\mathbf{W}_{D}^{L}(z_i-z_j)+\mathbf{W}_{D}^{L}\mathbf{W}(\sum_{v\in \mathcal{C}_j } \mathbf{h}_{v} ^{L-1}-\sum_{u\in \mathcal{C}_i } \mathbf{h}_{u} ^{L-1})\Big)\|\\
& \overset{\text{Lipschitz } \sigma} \leq
\|\mathbf{W}_{D}\|\|z_i-z_j\|+\|\mathbf{W}^{*}\|\|\Delta\|,
\end{align} where $\Delta = \sum_{v\in \mathcal{C}_j } \mathbf{h}_{v} ^{L-1}-\sum_{u\in \mathcal{C}_i } \mathbf{h}_{u} ^{L-1}$. We also assume that $\sigma$ is a subadditive function, such as ReLU.

Combined with Equation 21, we could get the following:
\begin{align}
\left<\vz_i, \vz_j\right>- \left<\vz_i', \vz_j'\right> 
&\leq\Big(\frac{1}{2} \|\mathbf{W}_{D}^{L}\|^{2}-1\Big)\|\vz_i-\vz_j\|^{2}+\|\mathbf{W}_{D}^{L}\mathbf{W}\|^{2}\|\Delta\|^{2}.
\,
\end{align}